\RequirePackage{fix-cm}
\documentclass[twocolumn]{svjour3}          
\smartqed  
\usepackage{graphicx}
\usepackage{amssymb}
\usepackage{amsmath}
\usepackage{hyperref}

\journalname{Comp Stat}
\begin{document}

\title{Bayes in Wonderland! Predictive supervised classification inference hits unpredictability
\thanks{AA devised the study and developed the theory, AA and VK advanced the models and implemented the methods, JT provided the funding, VK, JT, and AA wrote the paper. }
}

\titlerunning{Bayes in Wonderland!}        

\author{Ali Amiryousefi* \and
        Ville Kinnula \and
        Jing Tang 
}

\authorrunning{Kinnula, V \textit{et. al.}} 

\institute{Research Program in Systems Oncology,  \at
Faculty of Medicine, University of Helsinki, Finland 
               \\
              \\
              *Corresponding author. \email{ali.amiryousefi@helsinki.fi}           
           \and
}

\date{Received: date / Accepted: date}

\maketitle

\begin{abstract} 
The marginal Bayesian predictive classifiers (mBpc) as opposed to the simultaneous Bayesian predictive classifiers (sBpc), handle each data separately and hence tacitly assumes the independence of the observations. However, due to saturation in learning of generative model parameters, the adverse effect of this false assumption on the accuracy of mBpc tends to wear out in face of increasing amount of training data; guaranteeing the convergence of these two classifiers under de Finetti type of exchangeability. This result however, is far from trivial for the sequences generated under Partition exchangeability (PE), where even umpteen amount of training data is not ruling out the possibility of an unobserved outcome (Wonderland!). We provide a computational scheme that allows the generation of the sequences under PE. Based on that, with controlled increase of the training data, we show the convergence of the sBpc and mBpc. This underlies the use of simpler yet computationally more efficient marginal classifiers instead of simultaneous. We also provide a parameter estimation of the generative model giving rise to the partition exchangeable sequence as well as a testing paradigm for the equality of this parameter across different samples. The package for Bayesian predictive supervised classifications, parameter estimation and hypothesis testing of the Ewens Sampling Formula generative model is deposited on CRAN as \href{https://cran.r-project.org/web/packages/PEkit/index.html}{PEkit package} and \href{https://github.com/AmiryousefiLab/PEkit}{https://github.com/AmiryousefiLab/PEkit}. \\
\subclass{MSC code1 \and MSC code2 \and more}
\end{abstract}

\section{Introduction}
\label{intro}

Under the broad realm of inductive inference, the goal of the supervised classification is to assign the test objects into $a\, priori$ defined number of classes learned from the training data \cite{Solomonoff}. One of the most applicable machinery that can optimally handle these scenarios is Bayesian which with a given prior information and accruing observed data, gradually enhances the precision of the inferred population’s parameters \cite{Hand}. We consider here the general supervised classification case where the sets of species observed for features are not closed $a\, priori$, leaving the probability of observing new species at any stage non-negative. The de Finetti type of exchangeability \cite{Berlow}, seems intractable in these cases. Nevertheless, one solution is to adhere to a form of partition exchangeability due to Kingman \cite{Kingman2}. Assuming this type of exchangeability for each competing class, the derivation here shows that given an infinite amount of data, the simultaneous and marginal predictive classifiers will converge asymptotically. This is congruent with the similar study under the de Finetti exchangeability by Corander \cite{Corander}. Due to the existence of marginal dependency between the data points, the simultaneous and marginal classifiers are not necessarily equal. On the other hand, their convergence is not intuitive due to the complication posed with  \textit{a  priori} unfixed set of observable species. Upon availability of umpteen amount of data however, the proof presented here justifies the replacement of the marginal classifiers with the computationally expensive simultaneous ones. 
\par
 The following section introduces partition exchangeability and a probability distribution related to it. Hypothesis tests for its parameter are also introduced. Finally, the predictive classifiers under partition exchangeability are derived and the algorithms used to implement the classifiers are presented along with classification results on simulated data sets from partition exchangeable distributions.

\section{Partition exchangeability}

Assume that number of species related to our feature is unfixed {\it a priori}. Upon availability of the vector of test labels $S$, under the ${\it partition}$ ${\it exchangeability}$ framework, we can deduce the sufficient statistic for each subset of data. To define this statistic, consider the assignment of arbitrary permutation of integers 1, $\ldots$, |$s_c$| to the items in $s_c$
where $n_c = |s_c|$ is the size of a given class $c$. Introducing the $I(\ldots)$ indicator function and $n_{cl} = \Sigma_{i \in s_c} I(x_i = l)$ as the frequency of items in class $c$ having value $l \in {\bf \chi}$ , then in terms of count in ${\bf x}^{(c)}$ one can write the sufficient
statistic as

\begin{equation}
    \rho_{ct} = \sum_{l=1}^\infty I(n_{cl}=t) ,
\end{equation}

The vector of sufficient statistic $\rho_c =(\rho_{ct})_{t=1}^{n_c}$ indicates a partition of the
integer $n_c$ such that $\rho_{ct}$ is the frequency of specific feature values that have been 
observed only $t$ times in class $c$ of test data. Given the above formulation \cite{Kingman}, 
the random partition is exchangeable if and only if  two different sequences having the
same vector of sufficient statistics have the same probabilities of occurring. According to Kingman's representation theorem 
\cite{Kingman2}, the probability distribution of the vector of sufficient statistics under partition exchangeability will follow the Poisson-Dirichlet($\psi$) (PD) distribution known also as the Ewens sampling formula \cite{Ewens},

\begin{gather}
\label{eqn:Ewens}
    p(\boldsymbol{\rho}|\psi) = \frac{n!}{\psi(\psi+1) \ldots (\psi + n - 1)}  \prod_{t=1}^n  \Bigg\{ {(\frac{\psi}{t})^{\rho_t} \frac{1}{\rho_t!}} \Bigg\}, 
\end{gather}
\begin{gather*}
    \; \forall \: \psi \in \mathbb{R}^+, \;\; \boldsymbol{\rho} \in \mathfrak{S}_\rho,
\end{gather*}
where,
\begin{equation*}
\begin{split}
    \mathfrak{S}_\rho = \Bigg\{ (\rho_1,\rho_2, \ldots \rho_n) \: | \: \sum_{i=1}^n  i\rho_i = n,& \: \rho_i \in \mathbb{N}_0, \: \\
    &i=1,2,\ldots n \Bigg\} .
\end{split}
\end{equation*}

For a comprehensive review of the Ewens sampling formula and its history and further applications, see \cite{Crane}.


\subsection{Parameter estimation}\label{param-estim}

The parameter $\psi$ in \eqref{eqn:Ewens} is called a dispersal parameter, as the higher its value is the more distinct species will likely be observed in a sample from this distribution. A Maximum Likelihood Estimate (MLE) for the parameter $\psi$ can be derived from a sample as demonstrated in \cite{Ewens}. The estimate turns out to be the root of the equation

\begin{equation}
\label{eqn:mle}
    \sum_{j=1}^n \frac{\psi}{\psi + j - 1} = \sum_{t=1}^n \rho_t . 
\end{equation}

The proof of this is provided in Appendix. The sum of sufficient statistics on the right side of the equation equals the observed number of distinct species in the sample. The left side of the equation equals, as shown in by Ewens in \cite{Ewens}, the expected number of distinct species observed given parameter $\psi$. Thus, the MLE of $\psi$ is that number for which the observed number of distinct species equals the expected number of species observed in a sample of size $n$. There is no closed form solution for the equation for arbitrary $n$, so the MLE $\hat{\psi}$ has to be numerically searched for. As the right side of the equation is a strictly increasing function when $\psi>0$, a binary search algorithm can find the root. In supervised classification, an estimate for $\psi$ can be calculated for each class in the training data.

\subsection{Hypothesis testing}

A Lagrange multiplier test as defined in \cite{Rao} can be used for statistical testing of a hypothesized parameter $\psi_0$ under the null hypothesis $H_0: \: \psi=\psi_0$ for a single sample. The test statistic is constructed as follows:

\begin{equation}
    S(\psi_0) = \frac{U(\psi_0)^2}{I(\psi_0)} ,
\end{equation}
where $U$ is the gradient of the log-likelihood $L(\psi)$, and $I$ is the Fisher information of the distribution. Under the null hypothesis the test statistic $S$ follows the ${\bf \chi_1^2}$-distribution. In the case of the PD distribution, these quantities become

\begin{equation}
\label{eqn:logderivative}
    U(\psi_0) = \sum_{i=1}^n \Bigg ( \frac{\rho_i}{\psi_0} - \frac{1}{\psi_0 + i -1} \Bigg )
\end{equation}

\begin{equation}
\label{eqn:fisherinformation}
    I(\psi_0) = \sum_{i=1}^n \Bigg ( \frac{1}{\psi_0(\psi_0 +i-1)} - \frac{1}{(\psi_0 + i -1)^2} \Bigg )
\end{equation}
The proof of this is provided in Appendix.\par

For a multiple-sample test to infer whether there is a statistically significant difference in the $\psi$ of each sample, we have a devised a Likelihood Ratio Test (LRT) \cite{Neyman}. The null hypothesis of the test is that there is no difference in the $\psi$ of each of $s$ samples, $H_0: \: \psi_1=\psi_2=\ldots=\psi_s$, and consequently the alternative hypothesis is the inequality of at least two $\psi$. The test statistic $\Lambda$ is constructed as follows:

\begin{equation}
    \label{Lambda}
    \Lambda = -2\:log\frac{sup\:\mathcal{L}(\theta_0)}{sup\:\mathcal{L}(\hat{\theta})} \xrightarrow[]{\mathcal{D}} \chi_d^2,
\end{equation}
where $\mathcal{L}(\theta_{0})$ is the likelihood function of the data given the model under the null-hypothesis, and $\mathcal{L(\hat{\theta})}$ is the unrestricted likelihood of the model. The $sup$ refers to \textit{supremum}, so the likelihood is evaluated at the MLE of the parameters. $\Lambda$ asymptotically converges in distribution to the $\chi_d^2$, where $d$ equals the difference in the amount of parameters between the models. When testing the $\psi$ of $s$ different samples from the PD-distribution with possibly different sample sizes $n_s$, the model under the null hypothesis has one shared $\psi$, while the unrestricted model has $s$ different dispersal parameters, ($\psi_1, \psi_2, \ldots, \psi_t$), so $d=t-1$. The likelihood $\mathcal{L}$ of multiple independent samples from the PD-distribution is a product of the density functions for the partitions $\rho$ of those samples, and under $H_1$ the MLE of $\psi$ for each of the samples is evaluated as in \eqref{eqn:mle} from each sample independently, as the other samples have no effect on the $\psi$ of a single sample. However, under the null-hypothesis, the samples share a common $\psi$, the MLE of which according to \ref{param-estim} would be estimated by solving,

\begin{equation}
        \sum_{i=1}^s \sum_{j=1}^{n_s} \frac{\psi}{\psi + j - 1} = \sum_{i=1}^s \sum_{t=1}^{n_s} \rho_{st} . 
\end{equation}

As the likelihood of multiple independent samples with identical $\psi$ under the null-hypothesis is just a product of their likelihoods, the derivative of the log-likelihood is just a sum of the derivatives of each sample's likelihood. Again, the $\psi$ has to be obtained as the root of the above equation, which can be found with a small modification to the same binary algorithm as the one we use to determine the MLE of a single PD-distribution. Having found these MLEs of the $\psi$ under both hypotheses, the likelihood ratio in the test statistic in equation \eqref{Lambda} can then be expressed as,

\begin{equation}
    \prod_{j=1}^s \mathcal{L}(\boldsymbol{\rho}_j|\hat{\psi}_j) \:/\: \prod_{j=1}^s \mathcal{L}(\boldsymbol{\rho}_j|\hat{\psi}_j) ,
\end{equation}
where the likelihood function $\mathcal{L}(\boldsymbol{\rho}_j|\hat{\psi}_j)$ is the likelihood function of the PD($\boldsymbol{\rho}_j|\psi$) distribution for the partition $\boldsymbol{\rho}$ of the $j$-th sample. As the restricted model in the numerator can never have a larger likelihood than the unrestricted likelihood in the denominator, and both are positive real numbers, this ratio is bounded between 0 and 1.

\subsection{A note on two-parameter PD}
A two-parameter formulation of the distribution presented in \cite{Pitman}. The added parameter in PD($\alpha, \psi$) is called the discount parameter. The role of the parameters is discussed in length in \cite{Crane}. In short, the $\alpha$ is defined to fall in the interval $[-1, 1]$. When it is positive, it increases the probability of observing new species in the future proportional to the amount of already discovered species,  while decreasing the probability of seeing the newly observed species again. When $\alpha$ is negative, the opposite is true and the number of new species to be discovered is bounded. The single parameter $PD(\psi)$ is the special case of the two parameter distribution with $\alpha$ set to 0, $PD(0,\psi)$. The two parameter distribution is not considered further here, as the estimation of the parameters becomes a daunting task compared to the simple one parameter formalization.

\subsection{Supervised classifiers}
Consider the set of $m$ available training items by $M$ and correspondingly the set of $n$ test items by $N$. For each item, we observe only one feature 1 that can take value from species set ${\bf \chi} = \{1, 2, . . . , r\}$. Note that each number in $\chi$ is represented with one species such that the first species observed is represented with integer 1, the second species is represented with integer 2, and so on. On the other hand, $r$ is not known ${\it priori}$ denoting the fact that we are uninformative about all of the species possible in our population. A training item $i \in M $ is characterized by a feature $z_i$ such that, $z_i \in \chi$. Similarly, we have for a test item $i \in N $ the feature $x_i$ such that, $x_i \in \chi$. Collections of the training and test data features are denoted by vectors ${\bf z}$ and ${\bf x}$, respectively. Furthermore consider that the training data are allocated into $k$ distinct classes and $T$ is a joint labeling of all the training items into these classes. Simultaneous supervised classification will assign labels to all the test data
in $N$ in a joint manner. We can consider partitioning of $N$ test elements into $k$ different classes similar to $T$ such that $S = (s_1 , \ldots , s_k ), s_c \subseteq N, c = 1, ..., k$ be the joint labeling of this partition. The $T$ and $S$ structures indicate a partition of the training and test feature vectors, such that ${\bf z}^{(c)}$ and ${\bf x}^{(c)}$ represent the subset of training and test items in class $c = 1, ..., k,$ respectively. The $S$ denote the space of possible simultaneous classifications for a given $N$ and so $S \in \mathbb{S}$.

\par
The predictive probability of observing a new feature value of species $j$ given a set of prior observations from the PD distribution is

\begin{equation}
    p(x_{n+1}= j|{\bf n}) = \frac{n_j}{N+\psi}
\end{equation}
where $n_j$ is the frequency of species $j$ in the observed set ${\bf n}$. However, if the value $j$ is of a previously unobserved species, the predictive probability is
\begin{equation} \label{unseen pred}
    p(x_{n+1}= j|{\bf n}) = \frac{\psi}{N+\psi}.
\end{equation}

The proof of this arises both from the mechanics of an urn model that \cite{Hoppe} showed to generate the Ewens sampling formula, as well as from \cite{Karlin}. The predictive probability of a previously unseen feature value is thus higher for a population with a larger parameter $\psi$ than it is for a population of equal size with a smaller $\psi$. \par

Now the product predictive distribution for all test data that is assumes to be ${\it i.i.d}$ for the marginal classifier under the partition exchangeability framework becomes:

\begin{equation}
\begin{split}
    p_M &= \prod_{c=1}^k \prod_{i:S_i \in c} p( x_i=l_i| {\bf z}^{(c)}, T^{(c)}, S_i=c ) \\
    &= \prod_{c=1}^k \prod_{i:S_i \in c} \Bigg({\frac{m_{i;cl}}{m_c + \hat{\psi}_c}}\Bigg)^{I(m_{i;cl} \neq 0)}  \times \\ 
    &\quad\Bigg({\frac{\hat{\psi}_c}{m_c + \hat{\psi}_c}}\Bigg)^{I(m_{i;cl} = 0)}.
\end{split}
\end{equation}

Note that under a maximum $a \: posteriori$ -rule with training data of equal size for each class, a newly observed feature value that has not been previously seen in the training data for any class, will always be classified to the class with the highest estimated dispersal parameter $\hat{\psi}$. However, each class still has a positive probability of including previously unseen values based on the observed variety of distinct values previously observed within the class.
\par
Analogously to the product predictive probability for the case of general exchangeability in \cite{Corander}, the product predictive distribution of the simultaneous classifier under partition exchangeability, fitting the learning algorithm that is described in a later chapter, is then defined as

\begin{equation}
\begin{split}
    p_S &= \prod_{c=1}^k \prod_{i:S_i \in c} p( x_i=l_i| {\bf z}^{(c)}, T^{(c)}, S_i=c ) \\
    &= \prod_{c=1}^k \prod_{i:S_i \in c} \Bigg({\frac{m_{i;cl} + n_{i;cl}}{m_c + n_{i;cl} + \hat{\psi}_c}}\Bigg)^{I(m_{i;cl} \neq 0)} \times \\ 
    & \quad\Bigg({\frac{\hat{\psi}_c}{m_c+ n_{i;cl} + \hat{\psi}_c}}\Bigg)^{I(m_{i;cl} = 0)}.
\end{split}
\end{equation}

An asymptotic relationship between these classifiers is immediately apparent in these predictive probabilities. As the amount of training data in each class $m_c$ increases, the impact of class-wise test data $n_c$ becomes negligible in comparison, and the difference in the predictive probabilities approaches zero asymptotically. As the classifiers are searching for classification structures $S$ that optimize the test data predictive probability given the training data, and the predictive probabilities converge asymptotically, the classifiers are searching for the same optimal labeling.
\par
However, the classifiers handle values unseen in the training data differently. This is the situation where $m_{i;cjl}=0$ in the predictive probabilities $p_M$ and $p_S$ The marginal classifier's predictive probability for the test data is always maximized by assigning such a value into the class with the highest $\hat{\psi}$. The simultaneous classifier, however, considers the assignment of other instances of an unseen value as well. This can lead to optimal classification structures where different instances of an unseen value are classified into different classes. Thus, the convergence of the test data predictive probabilities of the marginal and simultaneous classifiers is not certain in the presence of unseen values. In practice, though, as the amount of training data $m$ tends to infinity, the probability of observing new values from the PD($\psi$)-distribution presented in equation \eqref{unseen pred} tends to 0: $\psi/(\psi + m)\: \rightarrow\:0$ as $m\rightarrow\infty$. Unexpected values would be very rare and would have a minimal effect on the classifications made by the different classifiers.

\subsection{Algorithms for the predictive classifiers}

\begin{table*}[t!]
\label{Tab:01}
\caption{The item-wise 0-1 classification error for the marginal and simultaneous classifiers, as well as the 0-1 difference between the predicted labels between the two classifiers.}  
{\begin{tabular*}{\textwidth}{@{\extracolsep{\fill}}llll@{}}   \\
\hline\noalign{\smallskip} 
& $\hat{p}_M$ & $\hat{p}_S$ & $|\hat{p}_M - \hat{p}_S|$\\
\noalign{\smallskip}\hline\noalign{\smallskip}
$m = 2*10^6, n = 2000, k = 3,$ ${\bf \psi}=1, 2$ & 0.491  & 0.491 & 0\\
$m = 1000, n = 2000, k = 3,$ ${\bf \psi}=1, 10, 50$  & 0.3408 & 0.2823 & 0.0626\\
$m = 1*10^5, n = 2000, k = 3,$ ${\bf \psi}=1, 10, 50$ & 0.2768 & 0.2758 & 0.0010\\
$m = 1000, n = 2000, k = 5,$ ${\bf \psi}=1, 100, 1000, 5000, 10000$ & 0.7535 & 0.5865 & 0.946 \\
$m = 2*10^6, n = 2000, k = 5,$ ${\bf \psi}=1, 100, 1000, 5000, 10000$ & 0.434 & 0.364 & 0.115\\
\noalign{\smallskip}\hline
\end{tabular*}}{}
\end{table*}

In this chapter the learning algorithms defined in \cite{Corander} are described. The predictive probabilities used in  these algorithms are defined above up to an unknown normalizing constant that, however, can be omitted as it doesn't affect the probability maximization step in the classification.\par
The marginal classifier is computationally attractive, as each test data point is individually classified according to a maximum ${\it a \; posteriori}$ rule:

\begin{equation}
    \hat{p}_M :  \underset{c=1,\ldots,k}{arg max} \; p(S_i=c | {\bf z}^{(M)},x_i, T),
\end{equation}

For the simultaneous classification algorithm, define classification structure $S_c^{(i)}$ to be an identical classification structure to $S$ with the item $i$ reclassified to class $c$. The greedy deterministic algorithm is then defined as

\begin{equation}
    \hat{p}_S :  \underset{S \in \mathbb{S}}{arg max} \; p(S | {\bf x}^{(N)},{\bf z}^{(M)}, T),
\end{equation}

\begin{enumerate}
    \item Set an initial $S_0$ with the marginal classifier algorithm $\hat{p}_M$.
    \item Until $S$ remains unchanged between iteration, do for each test item $i \in N$:
    \begin{equation}
        \hat{p}_S :  \underset{S \in \{S_1^{(i)}, \ldots , S_k^{(i)} \}}{arg max} \; p(S | {\bf x}^{(N)},{\bf z}^{(M)}, T)
    \end{equation}
\end{enumerate}

The simultaneous classifier thus works the same as the marginal classifier, but for each test item the potential labeling of other test items is also taken into account.

\section{Numerical illustrations of classifier performance and convergence}

\begin{figure}[!ht]
    \centering
    \includegraphics[width=0.455\textwidth]{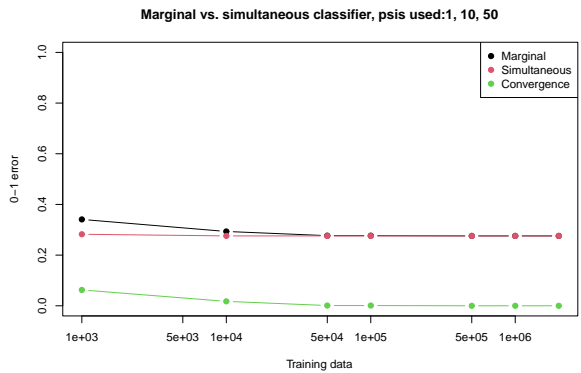}
    \caption{The classification error of the marginal and simultaneous classifiers with data sets from PD-distribution with $\psi\in (1, 10, 50)$, as well as the convergence of their labelings. Rows 2 and 3 in \ref{Tab:01} are included in the figure.}
    \label{fig:my_label}
\end{figure}

To study the classification performance and convergence of the two different classifiers, we simulated both training and test data sets from the PD distribution with a generative urn model described in \cite{Zabell} called the De Morgan process. We varied the amount of training and test data, the values of the distribution parameter $\psi$, as well as the amount of classes $k$. A large training data set of 4 millino data points was created for each $\psi$. A small sample of size 1000 was used to train the model first, and more of it was subsequently added until finally half of the whole data set was used as training data. A test data set created with the same parameters was kept constant for classification with all training data samples. An excerpt of the results can be found in Table 1, as well as Figure 1.\par
The table and the figures as well as the results not presented here show that the simultaneous classifier performs better than the marginal classifier especially as the amount of classes is higher and training data is small. As the amount of training data used increases, the predictions of the classifiers converge, as the extra information in the test data used by the simultaneous classifier becomes negligible. This behaviour is illustrated  in Figure 1. Additionally, the first row of the table shows that the classifier fails on the binary classification task with data sets created with $\psi$ of 1 and 2 respectively. This is because the classes created with $\psi$ of such similar magnitude will be distributed very similarly. Thus the classification results are no better than random guessing. The data is in fact so homogeneous in this case, that the labelings of the two different classifiers already converge with test data of size 1000. \par
In conclusion, the results support the hypothesis that the marginal and simultaneous classifiers converge in their labelings with enough training data. This justifies the use of the marginal classifier in place of the more accurate simultaneous one with large data. Still, the simultaneous classifiers is more accurate with smaller training data, as it benefits from the information in the test data.







\section{Discussion}
Previously unseen or unanticipated feature values are challenging scenarios in standard Bayesian inductive inference. Under general exchangeability as formalized by de Finetti, upon such an observation the entire alphabet of anticipated feature values must be retrospectively changed. The classifiers introduced in this article are, however, equipped to update their predictions autonomously based on a fully probabilistic framework. The classification of previously unseen feature values is handled through the use of a parameter learned from the training data, instead of for example using an uninformative Dirichlet-prior. 

The superiority in classification accuracy of labeling the test data simultaneously instead of one by one was also shown under partition exchangeability. The assumption that test data is $i.i.d.$ is obviously an unrealistic one. The two classifiers considered here only converge in prediction as the amount of training data approaches infinity. However, The computational cost of the simultaneous classifier increases exponentially as the amount of test data increases, making it unfeasible to use for large-scale prediction. Additionally, the algorithm presented here is only capable of arriving at local optima. Further research could be directed at the Gibbs sampler -assisted algorithm presented in \cite{Corander}, although the convergence of such algorithms with large test data sets is also uncertain. An implementation for the supervised classifiers considered here could also be devised for the two parameter distribution Poisson-Dirichlet($\alpha, \psi$).

%
%


\section*{Acknowledgements}

We thank ERC grant No. 716063 for the financial support.

\section*{Appendix}

\subsection*{\textbf{Maximum Likelihood Estimate}}
Here we provide the technical details for deriving the MLE in equation \eqref{eqn:mle} as well as the components $U$ and $I$ in the Lagrange Multiplier test described in equations \eqref{eqn:logderivative} and \eqref{eqn:fisherinformation}. The task is to find the first derivative of the logarithm of the Ewens sampling formula and its root in equation \eqref{eqn:Ewens}, as well as the second derivative needed for the Fisher information.

\begin{align*}
\begin{split}
L(\psi) &= log(n!) + \sum_{i=1}^n \{ -log(\psi + i -1) \,+ \\ 
&\rho_i\,log\,\psi - \rho_i\,log\,\rho_i - log(\rho_i!)  \} \\
=> l(\psi) &= \sum_{i=1}^n \{ -log(\psi + i -1) \,+ \rho_i\,log\,\psi \}
\end{split}
\end{align*}

\begin{align*}
\begin{split}
l'(\psi) &=\sum_{i=1}^n \Bigg( - \frac{1}{\psi + i -1} \,+ \frac{\rho_i}{\psi} \Bigg)\\
&= U(\psi) 
\end{split}
\end{align*}

The MLE is found by finding the root of the equation:

\begin{align*}
\begin{split}
\sum_{i=1}^n \Bigg( - \frac{1}{\hat{\psi} + i -1} \,+ \frac{\rho_i}{\hat{\psi}} \Bigg) = 0 \\
\sum_{j=1}^n \frac{\hat{\psi}}{\hat{\psi} + j + 1} = \sum_{t=1}^n \rho_t
\end{split}
\end{align*}

\subsection*{\textbf{Lagrange Multiplier Test}}

According to Ewens in \cite{Ewens}, the left side of the above equation equals the expected number of unique values observed with this $\psi$ and this sample size $n$, while the right side is the observed number of unique values:

\begin{align*}
    \begin{split}
        E[\sum_{t=1}^n \rho_t | \psi, n] = \sum_{i=1}^n   \frac{\hat{\psi}}{\hat{\psi} + i -1}
    \end{split}
\end{align*}

This is needed for the Fisher information $I$, along with the second derivative of $l(\psi)$:

\begin{align*}
\begin{split}
l''(\psi) &=\sum_{i=1}^n \Bigg( \frac{1}{(\psi + i -1)^2} \, - \frac{\rho_i}{\psi^2} \Bigg)\\
&= \sum_{i=1}^n  \frac{1}{(\psi + i -1)^2} -\frac{k}{\sum_{i=1}^n \psi^2},
\end{split}
\end{align*}

where $k$ denotes the observed amount of unique values in the sample. The Fisher information then becomes:

\begin{align*}
    \begin{split}
        I(\psi) &= -E[l''(k;\psi) | \psi] \\
        &= \frac{E[k|\psi]}{\sum_{i=1}^n \psi^2} - \sum_{i=1}^n  \frac{1}{(\psi + i -1)^2} \\
        &= \sum_{i=1}^n \Bigg(\frac{1}{\psi(\psi +i -1)} - \frac{1}{(\psi +i -1)^2} \Bigg),
    \end{split}
\end{align*}

where the expectation of $k$ is as described from above.

\end{document}